\PassOptionsToPackage{table,xcdraw}{xcolor}
\documentclass[square]{ws-procs11x85}

\usepackage{ws-procs-thm}           

\usepackage{makecell}    
\usepackage{subcaption}
\usepackage{wrapfig}

\usepackage[many]{tcolorbox}  
\usepackage{multirow} 
\usepackage{tabularx}

\definecolor{social_color}{HTML}{d7bee6}  
\definecolor{clinical_color}{HTML}{f5d496}  

\newtcbox{\modelbox}[1]{on line, colback=#1, colframe=#1, boxrule=0pt, arc=4pt, 
  left=3pt, right=3pt, top=1pt, bottom=1pt, boxsep=1pt}

\usepackage{booktabs}
\usepackage{adjustbox}

\begin{document}


\title{Inference Gap in Domain Expertise and Machine Intelligence in Named Entity Recognition: Creation of and Insights from a Substance Use-related Dataset}
\author{Sumon Kanti Dey,$^{1}$ Jeanne M. Powell,$^{1}$ Azra Ismail,$^{1}$ Jeanmarie Perrone,$^{2}$ Abeed Sarker$^{1,\dag}$}
\address{$^{1}$Department of Biomedical Informatics, Emory University, Atlanta, GA 30322, USA\\
$^{2}$Department of Emergency Medicine, Perelman School of Medicine, University of Pennsylvania, Philadelphia, PA 19104, USA\\
$^\dag$E-mail: abeed.sarker@emory.edu}




\begin{abstract}
Nonmedical opioid use is an urgent public health challenge, with far-reaching clinical and social consequences that are often underreported in traditional healthcare settings. Social media platforms, where individuals candidly share first-person experiences, offer a valuable yet underutilized source of insight into these impacts. In this study, we present a named entity recognition (NER) framework to extract two categories of self-reported consequences from social media narratives related to opioid use: \textit{ClinicalImpacts} (e.g., withdrawal, depression) and \textit{SocialImpacts} (e.g., job loss). To support this task, we introduce RedditImpacts 2.0, a high-quality dataset with refined annotation guidelines and a focus on first-person disclosures, addressing key limitations of prior work. We evaluate both fine-tuned encoder-based models and state-of-the-art large language models (LLMs) under zero- and few-shot in-context learning settings. Our fine-tuned DeBERTa-large model achieves a relaxed token-level F$_1$ of 0.61 [95\% CI: 0.43–0.62], consistently outperforming LLMs in precision, span accuracy, and adherence to task-specific guidelines. Furthermore, we show that strong NER performance can be achieved with substantially less labeled data, emphasizing the feasibility of deploying robust models in resource-limited settings. Our findings underscore the value of domain-specific fine-tuning for clinical NLP tasks and contribute to the responsible development of AI tools that may enhance addiction surveillance, improve interpretability, and support real-world healthcare decision-making. The best performing model, however, still significantly underperforms compared to inter-expert agreement (Cohen's kappa: 0.81), demonstrating that a gap persists between expert intelligence and current state-of-the-art NER/AI capabilities for tasks requiring deep domain knowledge. The dataset, annotation guidelines, appendix, and training scripts are publicly available to support future research.\footnote{\url {https://github.com/SumonKantiDey/Reddit\_Impacts\_NER}}
\end{abstract}


\copyrightinfo{\copyright\ 2024 The Authors. Open Access chapter published by World Scientific Publishing Company and distributed under the terms of the Creative Commons Attribution Non-Commercial (CC BY-NC) 4.0 License.}

\section{Introduction}
Nonmedical use of opioids remains a pressing public health challenge in the United States (U.S.), with more than 8.6 million Americans affected \cite{APA2023opioid}. Opioid-related overdoses have consistently remained a leading cause of accidental death in adults under 45 years of age, significantly reducing the average U.S. life expectancy \cite{MayoClinicOpioidPrevention}. In addition to fatal outcomes, nonmedical opioid use and addiction significantly disrupts social well-being and stability. People experiencing emotional distress or social instability, such as unemployment, housing insecurity, or family disruption, are more likely to initiate or escalate opioid use \cite{cerda2021critical, lin2024scoping}. Communities affected by the opioid epidemic often experience increased crime, educational challenges, and economic instability, further perpetuating cycles of disadvantage \cite{darolia2022social}.  

Nonmedical opioid use and subsequent impacts are underreported in clinical settings due to stigma, criminalization, and distrust of healthcare systems \cite{strike2020illicit}, leading to negative health outcomes \cite{strike2020illicit, cooper2017stigma, cheetham2022impact}. Widespread underreporting creates significant blind spots in public health surveillance, impeding efforts to detect early warning signs and deliver timely interventions. In contrast, social media sites like Reddit provide a pseudonymous environment where people feel more comfortable disclosing sensitive information. Health-related disclosures can include patterns of substance use, clinical symptoms (e.g., withdrawal), overdose risk, and co-occurring mental health issues \cite{lokala2024detecting, sarker2020mining}. Users also describe social consequences rarely captured in electronic health records \cite{giorgi2023predicting}, such as strained relationships, family disruption, financial hardship, unemployment, and social isolation. Analyzing these clinical and social impacts of opioid use reported on social media is crucial, as research has shown that the frequency and the content of opioid-related discussions online can mirror official epidemiological trends and provide timely insights for public health surveillance and intervention \cite{carpenter2025social}. The enormity of the opioid crisis, which has ravaged the U.S. for almost three decades, requires innovative solutions \cite{Humphreys2022StanfordLancet}, and the relative underutilization of social media data, which contains timely information directly posted by people with lived experiences, remains an untapped opportunity.


Extracting nuanced clinical and social impacts from informal, user-generated content poses significant challenges for natural language processing (NLP) models. Such content is highly unstructured, context-dependent, and contains abbreviations and ambiguities. Understanding the content, thus, requires deep contextual knowledge (e.g., through medical expertise or lived experience), which generic NLP systems, including large language models (LLMs) typically lack. Posts frequently express subjective, emotionally charged experiences, making it difficult for models to reliably map them to predefined categories \cite{coppersmith2018natural}. There is a critical need to bridge the gap between expert-level domain-specific knowledge and NLP model capabilities in characterizing and extracting meaningful information from such user-generated content, so that surveillance systems can be deployed at scale to inform public health strategies, intervention planning, and, ultimately, reduce the burden of the substance-related overdose epidemic \cite{beyer2023opioid}.
As an initial exploration of this challenge, our lab introduced a named entity recognition (NER) dataset---Reddit-Impacts \cite{ge2024reddit}---which was the first to capture both clinical and social dimensions of substance use. Although promising, efforts employing transformer-based models and proprietary LLMs (GPT-3.5) revealed several critical limitations in the annotation and subsequent NLP system performances. A more detailed discussion about these limitations is provided in Appendix~\ref {sec:limitation_of_the_previous_work}. To address these shortcomings and enable the development/training of more effective NLP systems, we update the data set and develop an improved processing pipeline to automatically identify the clinical and social impacts of non-medical opioid use in Reddit narratives. Our contributions are summarized below:

\begin{arabiclist}
\item We release \textbf{RedditImpacts 2.0}, an improved, task-specific dataset featuring detailed annotation guidelines, consistent entity spans, and exclusive focus on first-person narratives.
\item We propose an encoder-based framework for accurately extracting impact-related entities from unstructured social media narratives and systematically evaluate the effectiveness of various LLMs under zero- and few-shot in-context learning (ICL) settings.
\item We introduce custom evaluation metrics designed to effectively measure the accuracy and reliability of models, ensuring they only identify self-reported social and clinical impacts.
\smallskip
\item We conduct a targeted error analysis and data efficiency evaluation, demonstrating that strong, scalable performance can be achieved with substantially reduced labeled data, supporting responsible deployment in resource-limited and underserved settings.

\end{arabiclist}

Collectively, these contributions advance the development of human-aligned and trustworthy NLP systems to accurately interpret first-person opioid use narratives from social media. 

\textbf{Findings:} Our findings reveal that LLMs underperform in the token-level NER task to identify clinical and social impacts of nonmedical opioid use in social media posts, whereas encoder-based models can be fine-tuned to achieve substantially better performance. Via targeted error analysis, we highlight areas where fine-tuned encoders and LLMs differ, illustrating specific strengths and pitfalls of each. We also show that strong model performance can be achieved with significantly small training data, emphasizing the feasibility, scalability, and responsible development of fine-tuned NER models in low-resource and data settings.
\section{Related Work}
NER involves identifying specific terms or phrases within texts, referred to as ``entities". Traditionally, NER methods have concentrated on identifying narrow sets of predefined entities. In the biomedical field, for instance, entities include genes, diseases, chemicals, and proteins \cite{cho2019biomedical}. Compared to general or open-domain NER tasks, biomedical NER studies have received comparatively less scholarly attention because of the paucity of labeled data, the need for specialized computational models, and gaps in evaluation and benchmarking standards \cite{dai2021recognising, luo2023aioner, liu2024evaluating}. Biomedical NER involving social media data adds an additional layer of difficulty. Texts from such sources are characteristically informal, noisy, and linguistically diverse, making automated NER particularly difficult \cite{karisani2018did, magge2021overview}. Factors such as non-standard terminology, abbreviations, lack of context, and misspellings further amplify NER challenges \cite{magge2021overview, weissenbacher2019overview}.



Recent advances have overcome many of the abovementioned challenges, demonstrating the successful use of advanced NER models, including neural network and transformer-based approaches, to extract entities such as drugs, diseases, and symptoms from posts on platforms like  Reddit and online health forums \cite{scepanovic2020extracting, ge2024reddit, sidorov2025opioid}. Within the broader substance use sphere, transformer-based language models (BERT-base-NER, RoBERTa-base, BioBERT-base-cased, and Bio\_ClinicalBERT) have been utilized to accurately identify opioid-related entities \cite{sidorov2025opioid}. Scepanovic et al. \cite{scepanovic2020extracting} demonstrated that transformer-based models outperform traditional BiLSTM-CRF architectures in accurately extracting diverse medical entities such as symptoms, diseases, and drug names from social media posts, including Reddit. 


The emergence of instruction-tuned LLMs, including Llama 3 \cite{grattafiori2024llama}, Gemma 3 \cite{team2025gemma}, and GPT-4o \cite{hurst2024gpt}, which have demonstrated exceptional performance in medical reasoning benchmarks, have also, in many cases, led to improvements in challenging NER tasks primarily through innovative prompting strategies \cite{Zhan2025RAMIE,zhou2023universalner, shyr2024identifying}. These prompting techniques enable models to generalize effectively to previously unseen scenarios with minimal contextual cues, reducing the need for extensive fine-tuning and thus facilitating efficient biomedical entity extraction. There is, however, still a research gap in NER for highly imbalanced datasets, descriptive entities, and in low-shot settings, as demonstrated by past work on the Reddit-Impacts dataset \cite{ge2024reddit, Ge2023FewShot}. 

\vspace{-8pt}
\section{Methodology}
\begin{figure}[h]
    \centering
    \caption*{(a) Encoder-Based Pipeline for Impact Entity Recognition }
    \includegraphics[width=0.4\textwidth]{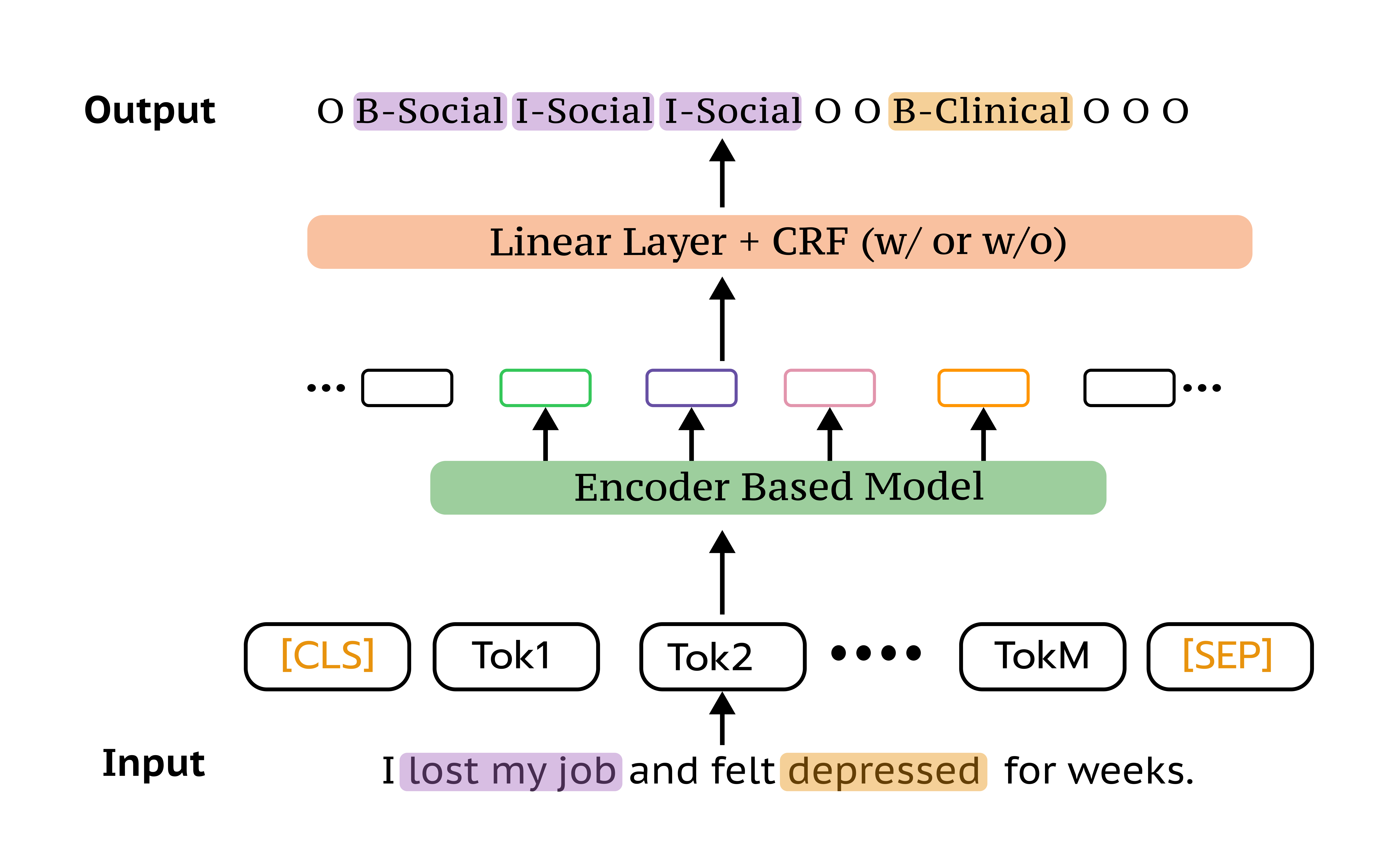}

    
    \caption*{(b) LLM-Based Pipeline for Impact Entity Recognition}
    \includegraphics[width=0.8\textwidth]{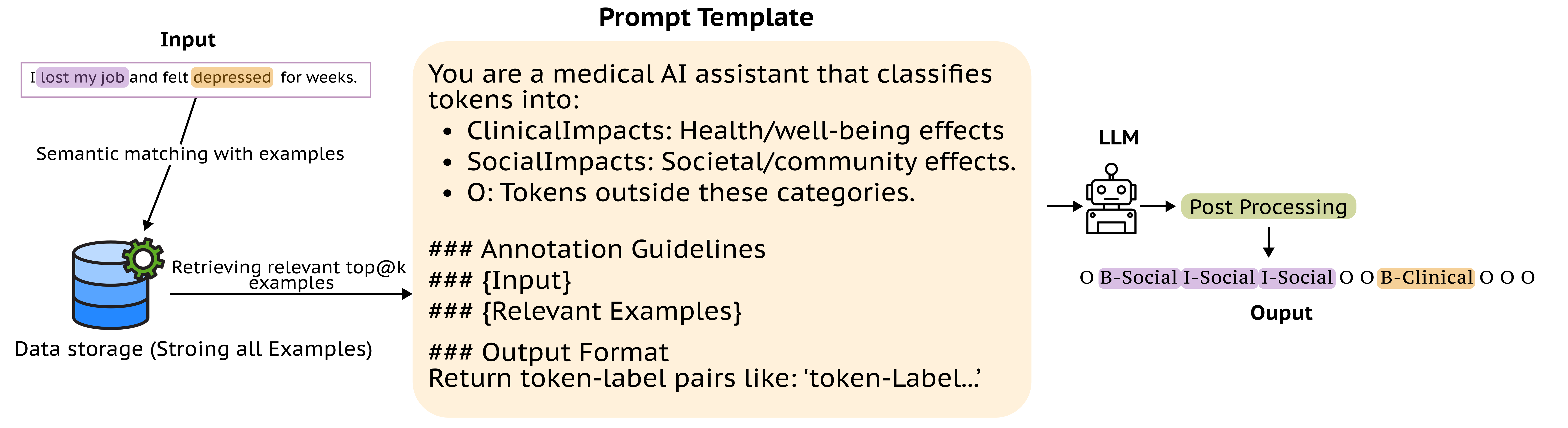}
    \vspace{0.3em}
    \caption{NER framework for detecting \textit{Social} and \textit{Clinical Impacts} in user-generated text. In the output, 
    \modelbox{clinical_color}{Clinical} corresponds to \modelbox{clinical_color}{ClinicalImpacts} and \modelbox{social_color}{Social} corresponds to \modelbox{social_color}{SocialImpacts}}.
    \label{fig:ner_vertical_comparison}
\end{figure}
\vspace{-2em}
\vspace{-12pt}


\subsection{Task Overview}

We formulate the NER task as a sequence-labeling problem using the BIO (Begin, Inside, Outside) tagging scheme. This scheme assigns one of three tags to each token: \texttt{B} to indicate the beginning of an entity, \texttt{I} for tokens that are inside an entity, and \texttt{O} for tokens that lie outside any entity span. Each tag is further associated with a specific entity type \texttt{X}, resulting in labels such as \texttt{B-X} (beginning of entity type \texttt{X}) and \texttt{O} (non-entity token). 

Given a token sequence, $T = [t_1, t_2, \ldots, t_n]$
the model predicts a corresponding sequence of labels $L = [l_1, l_2, \ldots, l_n]$, where each \( l_i \in \{\texttt{O}, \texttt{B-X}, \texttt{I-X} \} \). This formulation enables the model to learn both the boundary and the type of each entity span. In our study, we focus on two domain-specific entity types: \textbf{ClinicalImpacts}, representing physical or psychological consequences of substance use (e.g., \textit{``hospitalized"}, \textit{``depression"}), and \textbf{SocialImpacts}, which denote social, occupational, or relational consequences (e.g., \textit{``lost job"}, \textit{``arrested"}). An illustration of a BIO-tagged example with both entity types is shown in Table ~\ref{tab:bio-example}. 


\begin{table}[ht]
\centering
\caption{Example of BIO-tagged tokens. The phrase \textit{``lost my job''} is annotated as a \texttt{SocialImpacts} entity and \textit{``depressed''} as a \texttt{ClinicalImpacts} entity.}
\renewcommand{\arraystretch}{1.5}
\rowcolors{2}{gray!10}{white}
\begin{adjustbox}{width=\textwidth}
\begin{tabular}{l|ccccccccccc}
\toprule
\rowcolor{gray!25}
\textbf{Tokens} & I & lost & my & job & and & felt & depressed & for & weeks & . \\
\midrule
\textbf{BIO Tag} 
& O 
& \textbf{B-SocialImpacts} 
& \textbf{I-SocialImpacts} 
& \textbf{I-SocialImpacts} 
& O 
& O 
& \textbf{B-ClinicalImpacts} 
& O 
& O 
& O \\
\bottomrule
\end{tabular}
\end{adjustbox}
\label{tab:bio-example}
\end{table}

\subsection{Data Annotation}
Named entities for very specialized problems are expected to be context-dependent, making annotation of such entities inherently complex \cite{ding2021few}. This challenge is exacerbated in the context of social media data, where language is informal, unstructured, and shaped by personal, lived experiences. Posts related to opioid misuse often contain emotionally expressive narratives, fragmented grammar, and colloquial terms, making it difficult to consistently identify clinical and social impacts. To meet these complexities, this study included two experienced annotators with formal linguistic training. Both were guided by a comprehensive and detailed annotation manual, refined by a subject matter expert, specially designed to address the specific complexities of clinical and social impact mentions in opioid-related narratives. The full annotation guideline is provided in Appendix~\ref{sec:annotation_guidelines}.

The annotation process followed an iterative design. Initially, a 10\% subset of the dataset was independently annotated by both annotators. This step was critical for aligning their interpretation of the guidelines, especially in handling variations in language and subtle distinctions in meaning. Discrepancies were resolved through discussion, leading to refinement of the guideline. We conducted iterative co-annotation until an agreement accuracy exceeding 95\% was achieved. Once this threshold was met, the remaining data were divided between the two annotators for independent annotation.

We used Cohen’s Kappa \cite{cohen1960coefficient} across overlapping annotated samples to compute inter-annotator agreement. The resulting score of 81\% indicated substantial agreement, reflecting almost-perfect agreement \cite{VieraGarrett2005Kappa}. Descriptive statistics of the annotated dataset, including the total number of posts, tokens, and labeled entities, are summarized in Table~\ref{tab:reddit_stats}.



\begin{table}[ht]
\centering
\caption{Statistics of the annotated Reddit-Impacts NER dataset.}
\resizebox{0.9\textwidth}{!}{
\begin{tabular}{lccccc}
\toprule
\textbf{Split} & \textbf{Total Posts} & \textbf{SocialImpacts Entities} & \textbf{ClinicalImpacts Entities} & \textbf{Total Entities} & \textbf{Total Tokens} \\
\midrule
Train       & 842  & 408 & 616 & 1024 & 17.2K \\
Dev         & 258  & 167 & 223 & 390  & 5.2K  \\
Test        & 278  & 256 & 108 & 364  & 6.2K  \\
\textbf{Total} & 1378 & 831 & 947 & 1778 & 28.6K \\
\bottomrule
\end{tabular}}
\label{tab:reddit_stats}
\end{table}
\vspace{-1em}

\subsection{Experimental Setup}
For training our models, we merged the original training and development datasets to ensure we utilized the maximum amount of available data. We then set aside approximately 10\% of this combined dataset as a validation set, resulting in roughly a 90/10 split between training and validation. The test set was kept separate to provide a reliable assessment of how well our models perform on completely unseen data. To support fine-tuning these models, we leveraged a GPU equipped with 48GB of memory (e.g., NVIDIA RTX A6000). Detailed fine-tuning procedures and settings are provided in Appendix~\ref{sec:language_model_fine_tuning}.

\subsection{Modeling Approaches}
\label{sec:modeling_approach}
We evaluate three modeling paradigms for our SocialImpacts and ClinicalImpacts detection NER task: fine-tuning pre-trained language models (PLMs), augmenting PLMs with conditional random fields (CRF), and applying few-shot prompting techniques with large language models (LLMs). Each of these strategies has been extensively validated in recent NER research, motivating their adoption in our study \cite{lu2025large, keraghel2024survey, wang2023gpt, obeidat2025llms}. Figure~\ref{fig:ner_vertical_comparison} provides a visual summary of our overall modeling pipeline. 
\vspace{-12pt}
\subsubsection{Pre-trained language models (PLMs)}
We fine-tuned several transformer-based encoder derivatives, namely BERT \cite{devlin2019bert}, RoBERTa \cite{liu2019roberta}, DeBERTa \cite{he2020deberta}, RoBERTaNER \footnote{https://huggingface.co/Jean-Baptiste/roberta-large-ner-english}, and BioBERT \cite{lee2020biobert}. We replace each model's standard classification head with a linear token-level classification layer predicting our BIO-formatted labels. Models are optimized using cross-entropy loss, AdamW optimizer, learning rate scheduling with linear warm-up, and early stopping based on validation-set F$_1$ score.
\vspace{-12pt}
\subsubsection{PLM with CRF (PLM + CRF)}
Motivated by prior research demonstrating that adding CRF (Appendix~\ref{sec:crf}) layers significantly improves NER performance by capturing inter-label dependencies and ensuring valid BIO-tag transitions \cite{dang2018d3ner,qiu2025diffusion}, we extend each PLM by adding a CRF layer. The CRF replaces the softmax outputs and decodes the most probable sequence of labels through Viterbi decoding. Training is performed by minimizing the negative log-likelihood of the true sequences, further improving tag consistency and producing more coherent spans.
\vspace{-12pt}
\subsubsection{Few-Shot Prompting with LLMs}
We explore zero- and few-shot prompting strategies. For each input sequence requiring labeling, we perform three key steps. First, we compute the semantic embedding of all training samples using sentence-BERT \cite{reimers2019sentence}. Second, we identify and select the top@K most semantically similar training examples based on cosine similarity to the input sequence. We then construct a comprehensive prompt, integrating a concise task description, explicit annotation guidelines (BIO scheme and definitions for ClinicalImpacts and SocialImpacts), the top@K dynamically selected exemplar token-label sequences, and the target input tokens to be annotated. This in-context learning approach leverages the robust semantic understanding and few-shot inference capabilities of LLMs, eliminating the need for task-specific parameter updates and providing a flexible inference method suited for low-resource or rapid-deployment scenarios. The prompt is detailed in Appendix~\ref{sec:llm_prompt_details}

\subsection{Evaluation Metrics: Relaxed F$_1$ Score}

We evaluate model performance using a relaxed, token-level F$_1$ scoring method designed for entity recognition tasks involving partial matches. Our approach is inspired by partial-matching techniques used in biomedical NLP tasks \cite{andrade2023comparative, segura2013semeval}. Unlike strict span-level matching, which considers only exact boundary matches, relaxed evaluation accounts for partial overlap between predicted and ground truth entity spans of the same type, capturing cases where predictions are approximately correct. This evaluation is especially suitable for informal, user-generated text where exact span boundaries may be difficult to capture. The specific evaluation strategy formulation is outlined below.


Let each labeled sequence consist of predicted and gold spans represented as:
\[
G = \{g_1, g_2, \ldots, g_m\}, \quad P = \{p_1, p_2, \ldots, p_n\},
\]
where each span \( g_i \) or \( p_j \) is defined as a triple \( (t, s, e) \), denoting the entity type \( t \), start token index \( s \), and end token index \( e \).

We define token-level overlap between two spans \( g_i \) and \( p_j \) of the same entity type as:
\[
\text{Overlap}(g_i, p_j) = \max(0, \min(e_g, e_p) - \max(s_g, s_p) + 1),
\]
where \( (s_g, e_g) \) and \( (s_p, e_p) \) are the token span boundaries of the gold and predicted entities, respectively.


For each entity type \( T \), we compute:
\begin{equation*}
\text{TP}_{T} = \sum_{(g, p) \in \mathcal{M}_T} \text{Overlap}(g, p);
\text{P}_{T} = \sum_{p_j \in P_T} (e_p - s_p + 1); 
\text{R}_{T} = \sum_{g_i \in G_T} (e_g - s_g + 1) 
\end{equation*}
where \( \mathcal{M}_T \) denotes the set of span pairs \( (g, p) \) of type \( T \) with non-zero token-level overlap.

Precision, recall, and F$_1$ score for entity type \( T \) are then defined as:
\[
\text{Precision}_T = \frac{\text{TP}_T}{\text{P}_T}, \quad
\text{Recall}_T = \frac{\text{TP}_T}{\text{R}_T}, \quad
\text{F1}_T = \frac{2 \cdot \text{Precision}_T \cdot \text{Recall}_T}{\text{Precision}_T + \text{Recall}_T}.
\]

We additionally compute a micro-averaged overall F$_1$ score by aggregating overlapping token counts across all entity types:
\[
\text{Overall Precision} = \frac{\sum_T \text{TP}_T}{\sum_T \text{P}_T}, \quad
\text{Overall Recall} = \frac{\sum_T \text{TP}_T}{\sum_T \text{R}_T}, \quad
\]
\[
\text{Overall F$_1$} = \frac{2 \cdot \text{Overall Precision} \cdot \text{Overall Recall}}{\text{Overall Precision} + \text{Overall Recall}}
\]

This relaxed overlap-based metric for NER is tailored to informal user-generated content, where exact boundary prediction is often difficult.

\subsection{Error Analysis}
To better understand the strengths and limitations of our models, we conducted a qualitative error analysis comparing the fine-tuned DeBERTa-large model (the best-performing model among fine-tuned PLMs) and GPT-4o with 3-shot prompting (the top-performing model among LLMs) on the task of extracting social and clinical impacts.

\section{Results}

Table~\ref{tab:model_performance} summarizes the overall token-level relaxed precision, recall, and F$_1$ scores with 95\% confidence intervals (CI) for all models described in Section~\ref{sec:modeling_approach}.

Among the fine-tuned PLMs (Table~\ref{tab:model_performance}(a)), \textbf{DeBERTa-large} achieved the highest overall performance, with an F$_1$ score of \textbf{0.61} (95\% CI: [0.43, 0.62]), precision \textbf{0.75}, and recall 0.52, demonstrating significant improvement over prior approaches reported on this difficult NER task. Incorporating a CRF layer yielded mixed results across models. For DeBERTa-large, the F$_1$ score was lower with CRF at 0.57 [0.43-0.59], with overlapping confidence intervals suggesting statistically insignificant differences. For BERT-large, adding CRF similarly reduced  F$_1$ from 0.56 [0.38–0.56] to 0.52 [0.36–0.52]. In contrast, RoBERTaNER-large with CRF achieved an F$_1$ of 0.56 [0.45–0.60], slightly higher than its no-CRF counterpart at 0.51 [0.39–0.53], reflecting modest improvements in recall. Overall, the effect of CRF was minimal and appeared to depend on the underlying architecture. BioBERT models generally achieved lower F$_1$ scores compared to other PLMs, reflecting possible limitations in transferring their specialized clinical language understanding to the informal language context of social media. 

\begin{table}[h!]

  \centering
  \caption{Token-level relaxed precision, recall, and F$_1$ scores with 95\% confidence intervals (CI) across two evaluation settings:  
\textbf{(a)} \textit{Fine-tuned PLMs}--evaluated with and without Conditional Random Fields (CRF).  
\textbf{(b)} \textit{In-Context Learning Performance of LLMs}--evaluated under zero-shot, 3-shot, and 5-shot settings using the most similar examples for prompting.  
The best-performing model is highlighted in \textbf{bold}, and the second-best is \underline{underlined}.
}

  \label{tab:vertical_split_results}
  
  \begin{subtable}[t]{\textwidth}
    \centering
    \caption{Fine-tuned Pretrained Language Models (PLMs).}
    \renewcommand{\arraystretch}{1}
   \small 
   \resizebox{.8\textwidth}{!}{
    \begin{tabular}{lcccc}
      \toprule
      Model & Precision & Recall & F$_1$ & 95\% CI \\
      \midrule
      \rowcolor{gray!5}
      BERT{\scriptsize(large-uncased)} & 0.65 & 0.49 & 0.56 & [0.38, 0.56] \\
      \rowcolor{gray!5}
      BERT{\scriptsize(large-uncased)}{\scriptsize~+ CRF} & 0.67 & 0.42 & 0.52 & [0.36, 0.52] \\
      \rowcolor{gray!15}
      RoBERTaNER{\scriptsize(large)} & 0.67 & 0.42 & 0.51 & [0.39, 0.53] \\
      \rowcolor{gray!15}
      RoBERTaNER{\scriptsize(large)}{\scriptsize~+ CRF} & 0.59 & \textbf{0.54} & 0.56 & [0.45, 0.60] \\

      \rowcolor{gray!5}
      DeBERTa{\scriptsize(large)} & \textbf{0.75} & \underline{0.52} & \textbf{0.61} & [0.43, 0.62] \\
      \rowcolor{gray!5}
      DeBERTa{\scriptsize(large)}{\scriptsize~+ CRF} & \underline{0.68} & 0.50 & \underline{0.57} & [0.43, 0.59] \\
      
      \rowcolor{gray!15}
      BioBERT{\scriptsize(large-cased)} & 0.60 & 0.39 & 0.47 & [0.34, 0.52] \\
      \rowcolor{gray!15}
      BioBERT{\scriptsize(large-cased)}{\scriptsize~+ CRF} & 0.63 & 0.42 & 0.51 & [0.35, 0.53] \\
      \bottomrule
    \end{tabular}
    }
  \end{subtable}
  
  \vspace{1em}

\begin{subtable}[t]{\textwidth}
  \centering
  \caption{In-Context Learning Performance of LLMs under Zero-, 3-, and 5-shot Settings.}
  \renewcommand{\arraystretch}{1} 
  \small 
  \resizebox{.8\textwidth}{!}{
    \begin{tabular}{llcccc}
    \toprule
    Model & Prompting & Precision & Recall & F$_1$ & 95\% CI \\
    \midrule
    \rowcolor{gray!15}       & 0-shot & 0.29 & 0.35 & 0.32 & [0.27, 0.40] \\
    \rowcolor{gray!15} Gemma 3-27b-it & 3-shot & 0.26 & \textbf{0.48} & 0.34 & [0.33, 0.45] \\
    \rowcolor{gray!15}       & 5-shot & 0.25 & \textbf{0.48} & 0.33 & [0.32, 0.46] \\
    
    \rowcolor{gray!5}             & 0-shot & 0.46 & 0.27 & 0.34 & [0.28, 0.37] \\
    \rowcolor{gray!5} Llama 3-70b instruct  & 3-shot & \textbf{0.47} & 0.37 & 0.41 & [0.35, 0.45] \\
    \rowcolor{gray!5}             & 5-shot & \underline{0.46} & 0.37 & 0.41 & [0.33, 0.48] \\
    
     \rowcolor{gray!15}        & 0-shot & 0.43 & 0.37 & 0.40 & [0.31, 0.43] \\
      \rowcolor{gray!15} GPT-4o & 3-shot & 0.42 & \underline{0.46} & \textbf{0.44} & [0.39, 0.51] \\
      \rowcolor{gray!15}       & 5-shot & 0.39 & \textbf{0.48} & \underline{0.43} & [0.37, 0.51] \\
    \bottomrule
  \end{tabular}
  }
\end{subtable}
\label{tab:model_performance}
\end{table}

\begin{table}[ht!]
\centering
\caption{Entity-specific token-level relaxed precision, recall, F$_1$-score with 95\% CI for the top-performing fine-tuned PLM (DeBERTa-large) and few-shot LLM (GPT-4o, 3-shot).}
\renewcommand{\arraystretch}{1.2}
\small
\resizebox{.9\textwidth}{!}{
    \begin{tabular}{l l cccc}
\toprule
\textbf{Model} & \textbf{Entity} & \textbf{Precision} & \textbf{Recall} & \textbf{F$_1$-score} & \textbf{95\% CI} \\
\midrule
DeBERTa{\scriptsize(large)} & SocialImpacts   & 0.63 & 0.41 & 0.50 & [0.25, 0.62] \\
                            & ClinicalImpacts & 0.80 & 0.56 & 0.66 & [0.45, 0.67] \\
\midrule
GPT-4o{\scriptsize(3-shot)}       & SocialImpacts   & 0.28 & 0.25 & 0.26 & [0.18, 0.34] \\
                     & ClinicalImpacts & 0.46 & 0.56 & 0.51 & [0.44, 0.60] \\
\bottomrule
\end{tabular}
}
\label{tab:token_level_performance}
\end{table}

Table~\ref{tab:model_performance}(b) presents the in-context learning performance of LLMs under zero-, 3-, and 5-shot prompting conditions. Among these LLMs, \textbf{GPT-4o} achieved the best overall few-shot prompting performance, with an F$_1$ score of \textbf{0.44} [95\% CI: 0.37, 0.51] in the 3-shot setting, demonstrating balanced precision (0.42) and recall (0.46). Llama 3-70b also performed competitively, reaching an F$_1$ score of 0.41 in the 3- and 5-shot settings. Gemma 3-27b, while lower in absolute performance, showed a clear benefit from in-context examples, improving from an F$_1$ score of 0.32 [0.27-0.40] in a zero-shot setting to 0.34 [0.33-0.45] in the 3-shot setting. 

These results collectively demonstrate that few-shot prompting using semantically similar examples enhances the performance of LLMs on token-level prediction tasks compared to zero-shot settings. Nevertheless, even the best-performing LLM (GPT4o, F$_1$ = 0.44) remained below the best fine-tuned encoder (DeBERTa-large, F$_1$=0.61), emphasizing the advantage of domain-specific fine-tuning for token-level NER tasks in this domain.

We also report entity-specific token label performance for the best-performing fine-tuned PLM (DeBERTa-large) and the top-performing few-shot LLM (GPT-4o with 3-shot prompting) in Table~\ref{tab:token_level_performance}. In these two combinations, DeBERTa-large yielded the strongest performance on both ClinicalImpacts (F$_1$ = 0.60) and SocialImpacts (F$_1$ = 0.50). GPT-4o (3-shot) performed comparably on ClinicalImpacts (F$_1$ = 0.51), but underperformed on SocialImpacts (F$_1$ = 0.26), indicating varying levels of difficulty across entity types.

\subsection{Qualitative Error Analysis Results}
We identified four major categories of errors: label confusion, missed implicit entities, false positives due to negation/context errors, and violations of annotation guidelines.

\textbf{Label confusion (Social vs. Clinical):} GPT-4o frequently struggled to differentiate between social and clinical contexts. For instance, in the sentence \textit{``When I was 21... on a therapeutic community''}, the model misclassified the phrase \textit{``therapeutic community''} as a SocialImpact, whereas it was correctly annotated as a ClinicalImpact.

\textbf{Missed implicit entities:} Both models exhibited limitations in capturing impacts that were implied rather than explicitly stated. In the sentence \textit{``I was shocked... when I told them about my addiction and that I was seeking/needed help''}, DeBERTa successfully identified the ClinicalImpact \textit{``addiction''}, but failed to detect the SocialImpact embedded in the phrase \textit{``seeking/needed help''}. GPT-4o similarly failed, incorrectly labeling isolated words and missing the intended SocialImpacts entirely. 

\textbf{False positives due to context or negation errors:} Both models occasionally produced false positives when failing to interpret negation or surrounding context accurately. In the sentence \textit{``I am a recovering heroin addict with no criminal record but...''}, both DeBERTa and GPT-4o incorrectly labeled the phrase \textit{``criminal record''} as a SocialImpact, despite the explicit negation (``no criminal record'').

\textbf{Guideline violations and overgeneralization errors:} GPT-4o sometimes failed to adhere to annotation guidelines, including the instruction to annotate only first-person experiences. For example, in the sentence \textit{``Helps with the restlessness and anxiety''}, the model labeled \textit{``restlessness''} and \textit{``anxiety''} as ClinicalImpacts, despite the absence of first-person framing. Moreover, GPT-4o exhibited overgeneralization, labeling ambiguous or emotionally charged terms, such as \textit{``blindsided''}, \textit{``pressuring''}, and \textit{``treatment''}, as ClinicalImpacts, even when these were contextually neutral or not linked to direct impacts. 
\vspace{-12pt}

\section{Discussion}
\subsection{Bridging Expert Knowledge and Model Intelligence Remains a Challenge for Complex NER}
While our NER models demonstrated improvement over past efforts on the same, challenging dataset, there remains a significant gap between human-level agreement and model performance. Systematically improving the annotation guideline improved IAA, but models failed to fully capture the nuances of first-person reporting of clinical and social impacts of substance use. While our study explored strategies to bridge this gap using the Reddit-Impacts dataset, it generalizes to other NER tasks where deep domain expertise is necessary and in problems where annotated examples are lexically diverse and low due to dataset characteristics. In the context of substance use surveillance from social media data, improving NER performance may lead to the timely detection of emerging impacts of the continuously evolving overdose epidemic in the U.S. Our work may also shed light on how social media content reflects and shapes public perceptions of substance use over time, helping identify high-risk groups and the types of content they engage with, thereby informing development of more targeted and effective prevention strategies. Automated impact detection can further support the development of real-time support systems that connect users with peer or professional support during moments of crisis, thus strengthening clinical decision-making and public health intervention. 

\subsection{Error Analysis}
Label confusion, prominent in GPT-4o, highlights its difficulty in distinguishing between structured clinical environments and broader social circumstances, particularly when terminologies overlap, although these subtle differences are evident to human experts. GPT-4o's overgeneralization errors also revealed its difficulty in handling subtle context-dependent biomedical terms. Model errors in detecting implicit entities revealed challenges in grasping context-dependent, nuanced language. Labeling errors in both models' ability to process negated constructs and assess contextual validity prior to labeling.

Overall, while GPT-4o offers impressive generalization through prompting, it tends to misinterpret nuanced or implied information, and is more prone to guideline violations and contextual errors. In contrast, the fine-tuned DeBERTa-large model demonstrated stronger alignment with domain-specific guidelines and span accuracy. It may be possible to employ ensembling approaches, such as adding a first-person report classifier prior to processing by an LLM, rather than a single-module pipeline to improve inference performance. 
\subsection{Impact of In-Context Learning in LLMs}

Our findings indicate that ICL offers marginal performance gains in certain cases. For example, using 3-shot prompting resulted in a modest improvement of less than 4\% in F$_1$ score across all LLMs compared to the zero-shot setting. However, when we increased the number of examples to 5, the overall performance declined by $\approx$1--2\%, suggesting that additional examples may introduce noise or confuse the model's decision-making process in the NER detection task. The differences were not statistically significant, as indicated by overlapping 95\%-CIs. Overall, we observe that while ICL may enhance performance to a limited extent, particularly when using well-selected and minimal examples, increasing the amount of text in the context window may overwhelm the model, leading to reduced inference performance.

\subsection{Sample Size Consideration for Fine-Tuning a Language Model}
\begin{wrapfigure}{r}{0.5\textwidth} 
    \centering
    \includegraphics[width=0.5\textwidth]{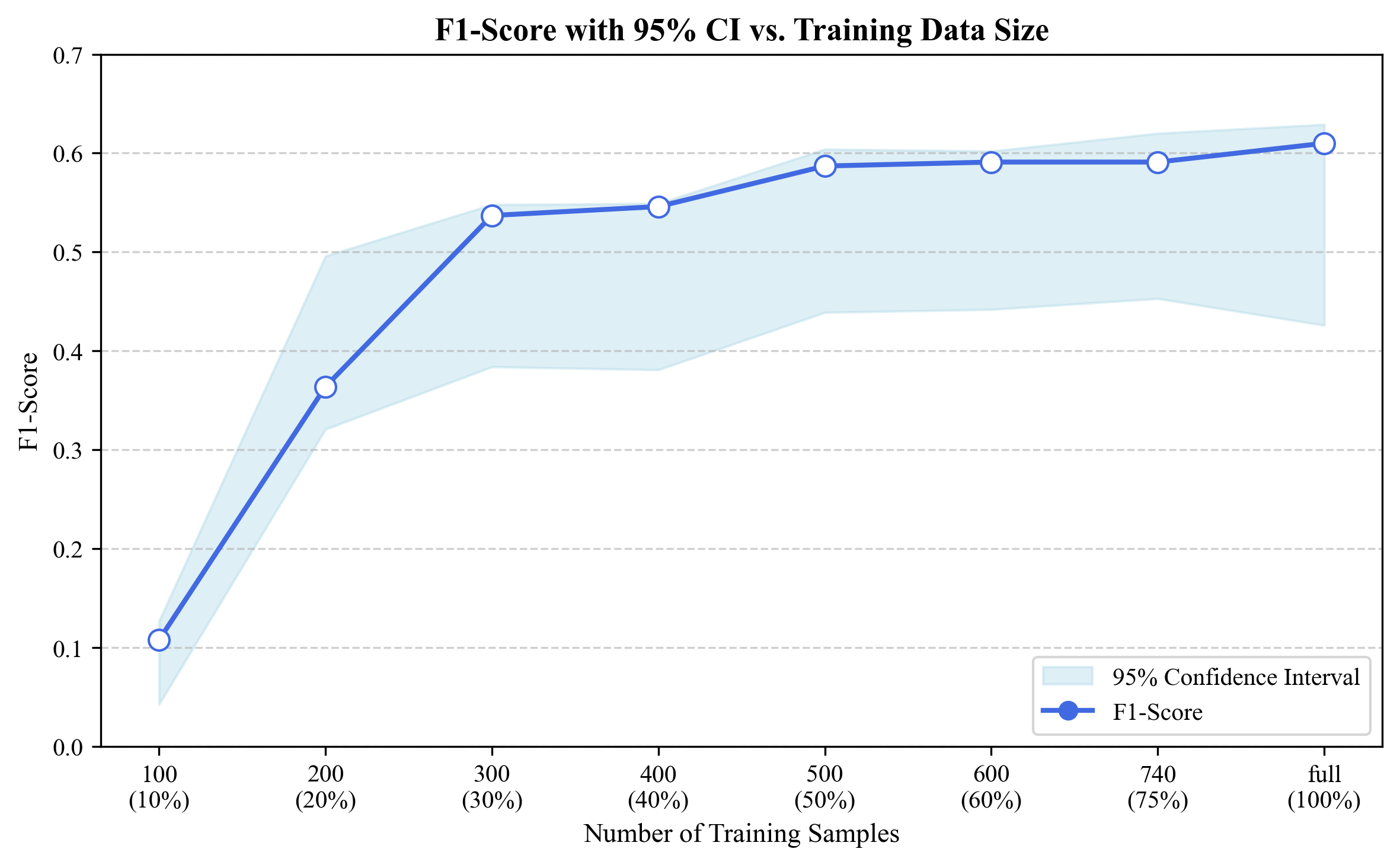} 
    \caption{F$_1$ score with 95\% CIs across training data sizes. x-axis shows the number of training samples with the corresponding percentage of the full dataset; shaded area represents the 95\% CI for each point.}
    \label{fig:learning_curve}
\end{wrapfigure}
Training a model for specialized tasks can be expensive due to the need for domain expert annotation---leading to substantial time and financial costs \cite{majdik2024sample, lopez2025clinical}. Research suggests that expert annotators take 10-30 seconds per sentence to label named entities \cite{chen2017active}. These challenges are further amplified in the case of user-generated biomedical data, which can be context-dependent, and emotionally nuanced, and where the demand for annotation quality is especially high. Consequently, developing models that can perform well even with limited annotated data is paramount.

We investigated how much annotated data is required to achieve expert-level performance using the encoder-based model, DeBERTa-large, which attained the highest F$_1$ score among all PLMs when trained on the full dataset. Our findings (Figure~\ref{fig:learning_curve}) demonstrate that F$_1$ score plateaus at approximately 50\% of training data, with additional data yielding only minor improvements. This suggests that better strategies for incorporating domain knowledge may be more effective in improving performance than annotating more data.





\subsection{Domain-Specific Fine-Tuning Still Matters for Biomedical NER}
Our findings reinforce the growing evidence that prompting alone is insufficient for achieving state-of-the-art performance in biomedical NER. Despite the flexibility and generalization capabilities of LLMs, models like GPT-4o and Llama 3-70B underperformed compared to domain-specific encoders fine-tuned on task-relevant labeled data. In our experiments, even the best-performing LLM using in-context learning (few-shot prompting) failed to match the F$_1$ score of the top fine-tuned encoder model, DeBERTa-large. These results align with some recent studies showing that LLMs, when used without task-specific fine-tuning, struggle to surpass traditional encoder architectures in specialized biomedical settings \cite{obeidat2025llms, lu2025large}. One underlying limitation is that token-level NER tasks---particularly in biomedical contexts---require models that can interpret the structured label schemes such as BIO tags and manage domain-specific terminology, which may include rare and unseen tokens. Prompting strategies, which do not modify the model's internal weights, are often insufficient to capture these nuances. In contrast, fine-tuning involves updating the model's parameters through supervised training, which allows the model to learn domain-specific representations and labeling conventions more effectively. The relatively stable and accurate performance of fine-tuned PLMs in structured biomedical NER tasks highlights the continued importance of fine-tuning, particularly in domains where general-purpose LLMs still fall short.
\subsection{Limitations}
We acknowledge some limitations in our study. First, the dataset used for training is relatively small, which is a common challenge when working with sensitive, first-person narratives describing substance use and its associated social and clinical impacts on social media. Expanding the dataset with more annotated examples may improve the robustness and generalizability of the model predictions. Second, while our fine-tuned encoder based models achieved strong performance, they are still struggle with capturing nuanced, context-dependent expression--especially when entities are implied rather than explicitly stated. Incorporating additional contextual signals or auxiliary tasks could further enhance the performance of the model. Lastly, though we evaluated LLMs under zero- and few-shot in-context learning settings, we did not explore instruction-tuned version of these models or feedback from error analysis to refine prompts. In future work, we aim to explore these approaches, along with multi-agent architectures, to better adapt LLMs for domain-specific entity recognition.
\section{Conclusion}
We explored strategies for extracting social and clinical impacts from first-person narratives related to substance use, using the refined and high-quality RedditImpacts 2.0 dataset. Our best-performing fine-tuned encoder model, DeBERTa-large, achieved a relaxed token-level F$_1$ score of 61\%, significantly outperforming the best-performing LLM (GPT-4o), which achieved an F$_1$ score of 44\%---a 17\% gap. This performance gap highlights the current limitations of prompting-based LLMs in specialized NER tasks without fine-tuning. Furthermore, our data efficiency analysis revealed that using only 50\% of the balanced dataset was sufficient to achieve performance comparable to training on the full dataset, suggesting that high-quality, domain-specific annotations, even in moderate quantities, can yield strong results.


\appendix{Limitation of the previous study}
Despite this importance and potential, relatively few studies have focused on the first-person experiences of individuals who misuse opioids. A notable effort in this area is the work by Ge et al. \cite{ge2024reddit}, which attempted to annotate and extract opioid-related impacts from Reddit posts. A notable effort in this area is the work by Ge et al. \cite{ge2024reddit}, which attempted to annotate and extract opioid-related impacts from Reddit posts. Upon closer examination, we identified multiple sources of annotation noise and inconsistency that limit the dataset’s utility for fine-grained, first-person impact detection. One major issue is the inclusion of entity spans from second-person and third-person perspectives. For example, the sentence ``You can taper as slow as possible and drop in the tiniest increments and still feel withdrawal because the alkaloid concentration varies so much \ldots'' was annotated as a clinical impact, despite being framed in the second person with no clear reference to the narrator’s own experience. Similarly, the sentence ``My mother struggled with mobility" was labeled as a clinical impact, although it describes a third-person experience. While such narratives may carry relevance in broader discourse analysis, our work is specifically concerned with identifying self-reported social and clinical impacts–––where the speaker directly references their own condition or circumstance. This distinction is particularly important for applications such as self-reported symptom tracking and personalized care modeling. 

We also found several cases of overgeneralized annotations where entire sentences were marked as impact entities, even though only a subset of words represented the actual impact. For instance, in the sentence ``So I did 45 days in a rehab here in Michigan,'' the word ``rehab'' represents the clinical impact, yet the full sentence was annotated as such. Similarly, ``I was homeless because of drugs'' was labeled entirely as a social impact, although only the word ``homeless'' denotes the impact. Another example includes the sentence ``I go to Alcoholics Anonymous, but I need more help/support,'' where only ``Alcoholics Anonymous'' should be identified as a social impact, but the whole sentence was tagged. In some instances, annotations were simply incorrect; for example, the sentence ``I had to wait like 38 hours, and it finally worked'' was labeled as a social impact, even though it does not convey any social or clinical consequence. Furthermore, neutral mentions of substances such as ``methadone,'' ``kratom'', ``Heroin'', and ``drugs'' were often annotated as clinical impacts, despite not indicating any actual impact or personal consequence in the context of those sentences. Annotation inconsistencies have been in Table \ref{tab:annotation-comparison}.

These issues introduce noise and reduce the effectiveness of machine learning models trained on the dataset, particularly transformer-based models, which have shown poor performance–––reportedly achieving zero F1 scores–––under these conditions.


\begin{table}[ht!]
    \centering
    \caption{Comparison of Original and Refined Annotations for Clinical and Social Impact Detection}
    \vspace{0.4em}
    \renewcommand{\arraystretch}{1.8}
    \begin{adjustbox}{width=\textwidth}
    \begin{tabular}{l|ccccccccccc}
        \toprule
        \rowcolor[HTML]{E6E6E6} 
        \textbf{Tokens} & I & was & a & homeless & junkie & on & the & streets & of & Florida & . \\
        \midrule
        \textbf{Original Annotation} & \multicolumn{8}{c}{\cellcolor[HTML]{FADBD8}SocialImpacts} & \_ & \_ & \_ \\
        \textbf{Refined Annotation} & \_ & \_ & \_ & \multicolumn{2}{c}{\cellcolor[HTML]{FCF3CF}SocialImpacts} & \_ & \_ & \_ & \_ & \_ & \_ \\
        \bottomrule
        \toprule
        \rowcolor[HTML]{E6E6E6} 
        \textbf{Tokens} & So & I & did & 45 & days & in & a & rehab & here & in & michigan . \\
        \midrule
        \textbf{Original Annotation} & \multicolumn{8}{c}{\cellcolor[HTML]{FADBD8}ClinicalImpacts} & \_ & \_ & \_ \\
        \textbf{Refined Annotation} & \_ & \_ & \_ & \_ & \_ & \_ & \_ & \cellcolor[HTML]{D4EFDF}ClinicalImpacts & \_ & \_ & \_ \\
        \bottomrule
    \end{tabular}
    \end{adjustbox}
    \label{tab:annotation-comparison}
\end{table}
\label{sec:limitation_of_the_previous_work}

\appendix{Conditional Random Fields (CRF)}
Conditional Random Fields (CRFs) \cite{lafferty2001conditional} are probabilistic models widely used for sequence labeling tasks such as Named Entity Recognition (NER). CRFs capture the dependencies between adjacent labels in a sequence, making them particularly effective in domains where contextual consistency is critical, such as social and clinical impact detection. Unlike models that assume independent label predictions, CRFs consider both the current token and its surrounding label structure, reducing inconsistent or invalid tagging patterns.

Let \( x = \langle x_1, x_2, \ldots, x_n \rangle \) be an input sequence of tokens and \( y = \langle y_1, y_2, \ldots, y_n \rangle \) be the corresponding sequence of labels. A linear-chain CRF defines the conditional probability of a label sequence given the input as:

\[
P(y \mid x) = \frac{1}{Z_x} \exp \left( \sum_{t=1}^{n} \sum_{k=1}^{K} \lambda_k f_k(y_{t-1}, y_t, x, t) \right)
\]

where \( Z_x \) is a normalization factor that sums over all possible label sequences, ensuring a valid probability distribution. Each \( f_k(y_{t-1}, y_t, x, t) \) is a feature function that encodes a specific combination of the previous label \( y_{t-1} \), the current label \( y_t \), the observation sequence \( x \), and the current position \( t \). These functions may represent transition features (e.g., from one label to another) or emission features (e.g., current word shape or embedding values). The parameters \( \lambda_k \) are learned weights that indicate the importance or reliability of each feature function during training. The label \( y_t \) corresponds to the class assigned to the word \( x_t \) at time step \( t \), while \( y_{t-1} \) represents the label of the preceding token.

By learning both emission and transition dynamics, CRFs reduce label ambiguity and enforce coherent tag sequences across the input. In this study, the CRF layer is applied on top of contextualized token embeddings derived from a transformer encoder, enabling robust recognition of entities related to social and clinical impacts.
\label{sec:crf}

\appendix{Annotation Guidelines}
We annotated entities belonging to two categories--\textit{SocialImpacts} and \textit{ClinicalImpacts}--in first-person narratives of opioid misuse. The following guidelines were applied to ensure consistency, reliability, and alignment with the intended task objectives.
\subsection*{General Instructions}
\begin{itemize}
    \item Annotations capture only meaningful, self-reported consequences of opioid misuse expressed by the individual.
    \item Both \textit{SocialImpacts} (e.g., job loss, family disruption) and \textit{ClinicalImpacts} (e.g., withdrawal, depression, hospitalization) are included.
    \item All annotations must reflect the individual's own lived experience.
\end{itemize}

\subsection*{Inclusion Criteria}
Entities were annotated when they met all of the following criteria:
\begin{itemize}
    \item \textbf{First-person account:} A social or clinical impact was annotated only if it was described in a first-person account directly related to the poster. \\
    \textit{Example:} ``I lost my job'' $\rightarrow$ annotated; ``My brother lost his job'' $\rightarrow$ not annotated.
    \vspace{0.3em}
    
    \item \textbf{Ambiguous context (assumed impact):} When opioid involvement could not be ruled out, the impact was assumed to be related. \\
    \textit{Example:} ``It caused me to fight with my family'', where ``fight with my family'' annotated as \textit{SociallImpacts}.
    \vspace{0.3em}
    
    \item \textbf{Polysubstance mention:} If a post mentioned multiple substances, annotate assuming opioid misuse contributed. \\
    \textit{Example:} ``I abuse alcohol and heroin, which has affected my health'' $\rightarrow$ annotated as \textit{ClinicalImpacts}.
    \vspace{0.3em}
    
    \item \textbf{Mental health symptoms:} Mental health issues were annotated as \textit{ClinicalImpacts} unless explicitly attributed to another cause. \\
    \textit{Included:} ``I feel depressed all the time.'' \\
    \textit{Excluded:} ``We broke up, so I am sad.'', where sad is clearly linked to the breakup and not opioid use 
    \vspace{0.3em}
    
    \item \textbf{Care-seeking behavior:} Mentions of rehab, counseling, or treatment were annotated as \textit{ClinicalImpacts}. \\
    \textit{Example:} ``I went to rehab last month'' $\rightarrow$ ``went to rehab'' is annotated.
\end{itemize}

\subsection*{Exclusion Criteria}
The following were explicitly excluded:
\begin{itemize}
    \item \textbf{Third-person accounts:} Impacts involving friends, family, or others. \\
    \textit{Example:} ``My brother lost his job'' $\rightarrow$ not annotated.
    \vspace{0.3em}
    
    \item \textbf{Drug names:} Mentions of specific drugs were not annotated as impacts (captured separately).
    \vspace{0.3em}

    \item \textbf{Personal pronouns in spans:} Personal pronouns (e.g., ``I'', ``he'', ``she'', ``they'', ``my'', ``our'') were excluded from the annotated span if they did not contribute directly to the social or clinical impact. \\
    \textit{Example:} ``I lost my job'' $\rightarrow$ span: \textit{lost my job}.
    \vspace{0.3em}

    \item \textbf{Modifiers in spans:} Temporal references, adjectives, adverbs, and similar modifiers were excluded unless integral to meaning. \\
    \textit{Example:} ``I am feeling really tired and crummy'' $\rightarrow$ span: \textit{tired and crummy}.
\end{itemize}
\label{sec:annotation_guidelines}

\appendix{Fine-tuning and LLM Prompting Details}
\subappendix{Language Model Fine-tuning}
We fine-tuned several encoder-based Pretrained Language Models (PLMs) on our annotated dataset, experimenting with various hyperparameters to identify optimal configurations for the Named Entity Recognition (NER) task. We explored multiple combinations of learning rates, batch sizes, and other critical hyperparameters, as well as configurations both with and without a Conditional Random Field (CRF) layer. Table~\ref{tab:finetune_params} summarizes the ranges of key hyperparameters examined during model training.

\begin{table}[ht]
\centering
\renewcommand{\arraystretch}{1.2} 
\caption{Fine-tuning hyperparameter ranges explored across multiple encoder-based pre-trained language models (PLMs), with and without a Conditional Random Field (CRF) layer.}
\vspace{0.4em}
\label{tab:finetune_params}
\begin{tabular}{ll}
\hline
\textbf{Hyperparameter} & \textbf{Values Explored} \\ \hline
Base Model                & Multiple PLMs (encoder-based) \\
Batch Size                & \{8, 16, 32\} \\
Learning Rate             & \{1e-5, 2e-5, 3e-5, 5e-5\} \\
Dropout                   & \{0.2, 0.3\} \\
Number of Epochs          & \{7,10\} \\
Weight Decay              & 0.01 \\
Learning Rate Scheduler   & Linear \\
Gradient Clipping (max grad norm) & 1.0 \\
Early Stopping Patience   & 3 \\
Conditional Random Field (CRF) & \{True, False\} \\
\hline
\end{tabular}
\end{table}
\label{sec:language_model_fine_tuning}

\subappendix{LLM Prompting Details}
For evaluating LLMs such as GPT-4o, LLaMA3-70B, and Gemini, we utilized the structured prompt template described in Table~\ref{tab:prompt_template}. To ensure consistency and reproducibility across all LLM-based experiments, we set the temperature to 0.2, emphasizing deterministic and reliable outputs.

\begin{table}[ht]
\centering
\small
\caption{Prompt used for in-context learning. Few-shot examples (\(n=3\) or \(n=5\)) are dynamically selected based on semantic similarity to the input.}
\label{tab:prompt_template}
\begin{tabular}{p{0.95\textwidth}}
\toprule
You are a medical AI assistant that classifies tokens in a Reddit post into the following categories:  \\[0.3em]
\textbf{ClinicalImpacts:} Health and well-being effects.  \\ 
\textbf{SocialImpacts:} Societal or community-level effects.  \\ 
\textbf{O:} Tokens outside these categories.  \\[0.5em]

\textbf{\#\# Strict Annotation Rules \#\#}  \\
1. Annotate \textbf{ONLY} first-person experiences. Ignore third-party reports.  \\
2. Label all drug names (e.g., \texttt{heroin}, \texttt{fentanyl}) as \texttt{O}.  \\
3. Label personal pronouns (e.g., \texttt{I}, \texttt{my}) as \texttt{O} -- they are not part of entity spans.  \\
4. ASSUME opioid involvement unless a non-opioid cause is clearly stated. \\
5. If multiple substances are mentioned, default to opioid-related impact when unsure.  \\
6. Label mental health terms (e.g., \texttt{depression}) as \texttt{ClinicalImpacts} unless context clearly shows a non-opioid cause.  \\
7. Label non-integral words (e.g., adjectives, adverbs, or temporal words like \texttt {very}, \texttt{suddenly}) as \texttt{O} if they are not essential to the entity span.  \\
8. Corrupted or unreadable tokens (e.g., \texttt{Ìm}, \texttt{?}, \texttt{\#\#}) must be labeled as \texttt{O}.  \\
9. Maintain the exact token order and label all tokens.  \\
10. If unsure about a token, label it as \texttt{O}.  \\[0.5em]

\textbf{\#\# Output Format \#\#}  \\
Return token-label pairs in the following format:  \\
\texttt{token-Label token-Label token-Label ...}  \\[0.5em]

\textbf{Few-shot Examples (Top-N, dynamically retrieved):}  \\
\textit{Example 1:} \{token-Label token-Label token-Label ...\} \\
\textbf{...}  \\
\textit{Example 3:} \{token-Label token-Label token-Label ...\} \\
\textbf{...}  \\[0.5em]

\textbf{New Input:}  \\
\texttt{Tokens: [token\_1, token\_2, ..., token\_n]}\\
\textbf{Output:}\\
\bottomrule
\end{tabular}
\end{table}
\label{sec:llm_prompt_details}

\pagebreak

\bibliographystyle{ws-procs11x85}
\bibliography{ws-pro-sample}
\end{document}